# TAPe+ML: A Compact Structured Representation for Multi-Task Computer Vision

*Sergey Kurinov, Alexey Upatov*
*Comexp Research Lab, TAPe + ML Project*
*Nizhniy Novgorod, Russia*

*Correspondence:* tape@comexp.net
*Project page:* [ml.comexp.net](ml.comexp.net)

## 1. Abstract (Updated)

We present **TAPe+ML v3.5**, a compact multi-task computer vision architecture based on the Theory of Active Perception (TAPe). TAPe (Theory of Active Perception) is a way of representing visual information in which an image is described not by raw pixel values but by structured elements that preserve the similarity and difference relationships that matter for recognition.

So unlike conventional scaling paradigms reliant on billion-parameter foundation models and GPU clusters, TAPe+ML encodes structured relations among perceptual elements prior to recognition using a unified core with fewer than 100,000 parameters and a complete deployment footprint of 4.2 MB.

In this updated release, we report substantial gains across object detection and instance segmentation, and report single-core CPU latency against YOLO26s and RF-DETR-S. On COCO val2017, TAPe+ML v3.5 achieves **71.1 box AP** (86.6 AP50) and **60.3 mask AP** (82.8 mask AP50), outperforming large-scale detectors such as DINOv3 + Plain-DETR (66.1 box AP) and YOLO26x (57.5 box AP) using less than 1/70,000 of the parameters. On RF100-VL, it reaches **69.7 box AP**.

Furthermore, TAPe+ML v3.5 achieves **88.0% top-1 accuracy** on ImageNet-1k, **79.5%** on ObjectNet, and **75.7%** on ImageNet-Sketch within the same shared backbone. Under strict single-core CPU constraints across seven processor configurations, the model processes frames in 8.6–34.7 ms (29–116 FPS), delivering a 27 to 120× speedup over YOLO26s and up to 477× over RF-DETR-S.

In related experiments, TAPe also proved effective for scene detection in video: indexing one hour of video took 10-11 seconds with an index size under 1 MB, and new tasks on the test bench were learned from dozens of images instead of thousands.

These results show that strong computer vision performance can be achieved not only by scaling up the number of parameters and the amount of pre-training, but also through a more informative primary representation of visual information. TAPe+ML v3.5 is therefore proposed not merely as another compact model, but as a step toward a unified recognition core across the image-classification, object-detection, and instance-segmentation settings evaluated in this study.

While the core sections of this report detail the foundational TAPe+ML v3.0 architecture, methodology, and baseline experiments, Section 8 presents the architectural refinements and comprehensive benchmark results for the latest v3.5 release.

# 2. Introduction

## Issue 1 – Zoo of Models

In industry, computer vision tasks are often handled by separate models: one backbone for classification, another for detection, and a third for segmentation. Each is trained on different data, has different input and output formats, and requires its own infrastructure. For a large company, this means maintaining 5-10 separate pipelines.

Models that can handle detection and segmentation with a single network (YOLO [1], RF-DETR [2]) still have significant drawbacks:

- The pipelines of these models are poorly connected to one another – training happens in parallel, so, for instance, the segmentation task does not "feed" detection, and vice versa;
- They require bloated backbones "warmed up" far beyond the scope of the downstream tasks. For example, RF-DETR's backbone is enormous and is pre-trained on the unrelated ImageNet [3] task purely to support the model's overall operation.

TAPe+ML replaces separate task-specific backbones with a single TAPe perception space and a common recognition core across the classification, detection, and segmentation settings reported in this work. ImageNet [3], COCO [4], LVIS [5], and other benchmark datasets are used to evaluate the same system under the established protocols for different recognition tasks; they do not imply separate large backbones or independent vision pipelines. This architectural design is based on TAPe filters, which encode relations among visual elements before the recognition stage.

## Issue 2 – The Discrete Pixel Trap

When a neural network operates on pixels, it is actually solving two tasks at once: (1) reconstructing image structure from essentially arbitrary numbers, and (2) solving the target task (detection, classification). This imposes a huge, unnecessary burden.

Pixel-based CNN [6] and transformer models [7] learn visual structure from image tensors through their learned feature-extraction mechanisms. TAPe instead provides a structured representation before the recognition stage. The present work investigates whether this representation can support compact models in the evaluated classification, detection, and segmentation settings.

TAPe “intercepts” the image before the pixel level, using T-bits—units of information that already carry a similarity structure. As a result, the model no longer needs to “guess” the connections between patches, since relevant relations are already embedded in the representation. This is precisely why we were able to abandon transformers in TAPe+ML: global attention was not required for the evaluated tasks once the relevant structure was available in the input.

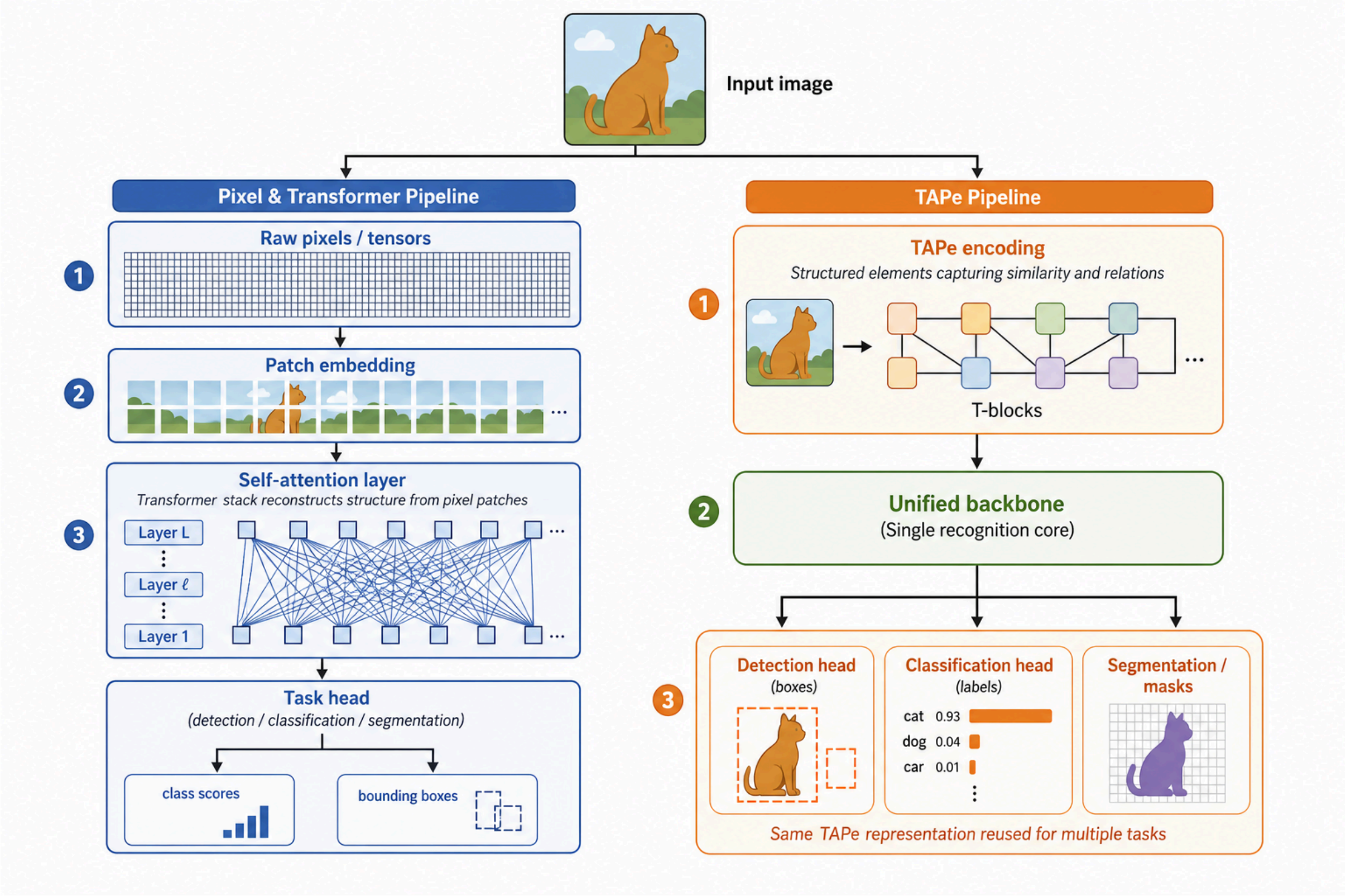

*Figure 1. TAPe+ML pipeline: structured TAPe representation as input to the shared recognition architecture*

## Issue 3 – Data and Annotation Cost

Meta’s DINOv2 [8] requires 142 million images because pixels are inherently unstable. In COCO evaluation, small objects are defined by an area below $32^2$ pixels [4]. A 30×50 pixel RGB region contains 1,500 pixels and, under 8-bit RGB encoding, has $256^{4500}$ possible raw-value configurations. Learning invariances from raw pixel inputs can require exposure to substantial visual variation. The model must "see" an enormous number of variations to learn stable rules.

Because the TAPe representation is structured, the model achieves the same stable rules from just 20 images for a new class and is fully retrained in under a minute. This is a fundamentally different order of magnitude: dozens of images instead of months of annotation.

**Accuracy graphs versus the number of images per class (20 → 50 → 200 → 500):**

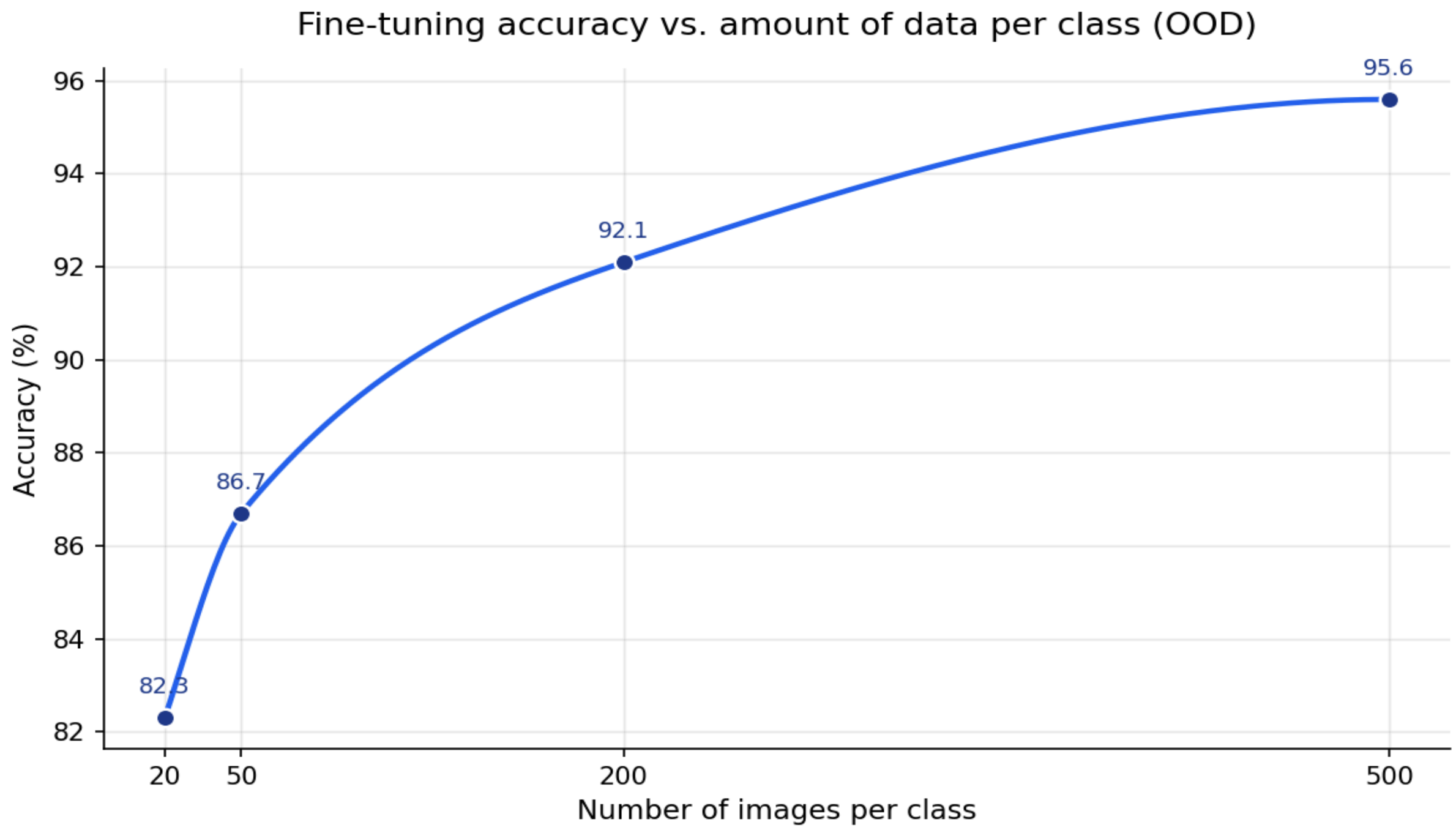

*Figure 2. Data-efficiency curve: model accuracy versus the number of labeled images per class (linear scale)*

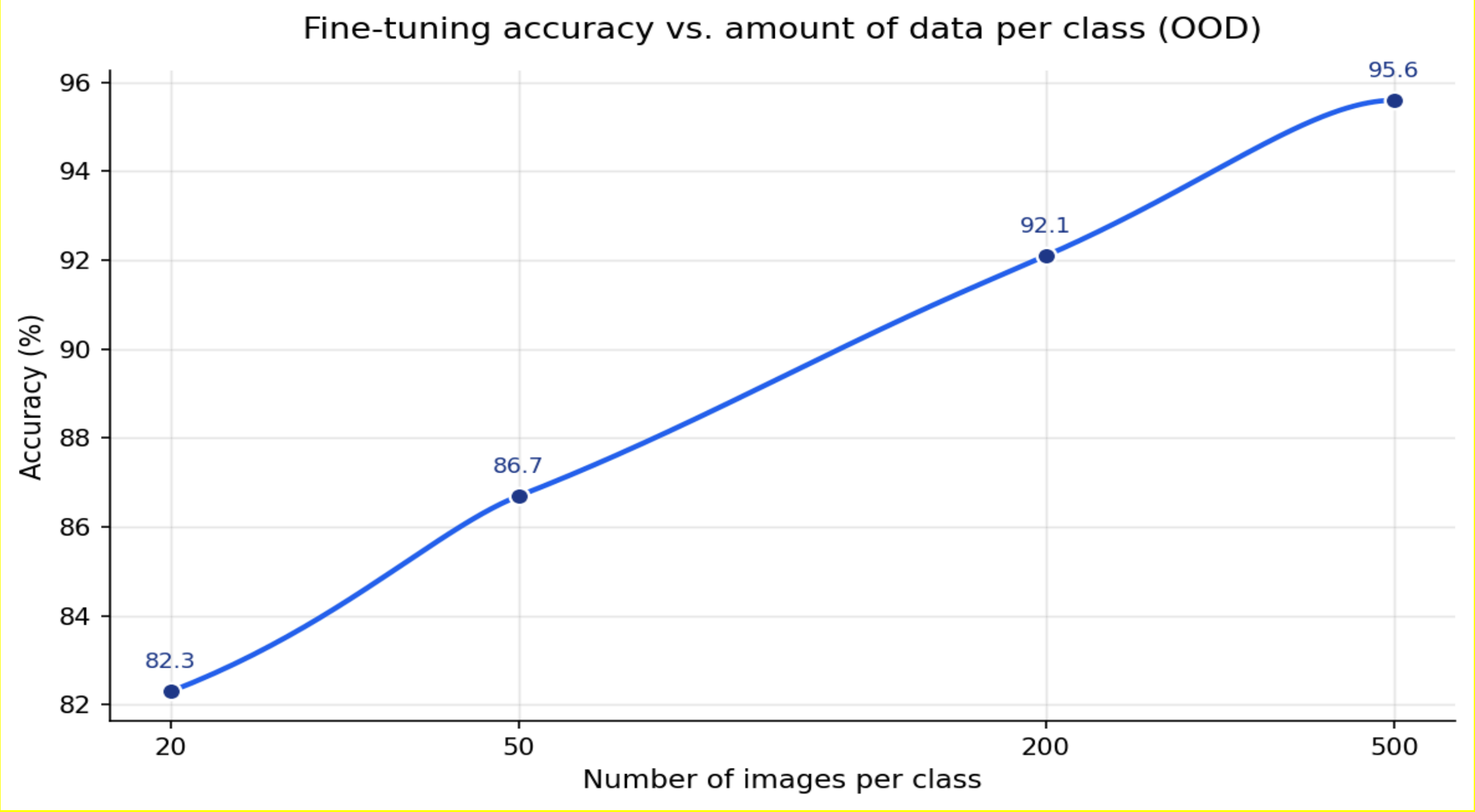

*Figure 3. Data-efficiency curve: model accuracy versus the number of labeled images per class (logarithmic scale).*

## Paper Organization and Release Scope

The remainder of this technical report is structured as follows. The core sections detail the theoretical foundations of TAPe, the shared recognition architecture, and the comprehensive baseline evaluations established in the **TAPe+ML v3.0** release across classification, detection, and segmentation. Finally, **Section 8** introduces the **TAPe+ML v3.5** release updates, presenting architectural refinements, expanded multi-task benchmarks (COCO, RF100-VL, ImageNet-1k, ObjectNet, and ImageNet-Sketch), and single-core CPU latency comparisons against contemporary foundation and real-time models.

# 3. Related Work

Modern computer vision research can be roughly grouped into several strands: methods that improve the representation of input data, general-purpose backbone architectures, specialized detectors and segmenters, and lightweight edge-oriented models. TAPe+ML intersects with all of these, but differs in combining them within a single system: a structured perception representation, a single backbone for multiple tasks, strong performance on standard benchmarks, and an extremely small parameter budget.

## 3.1. Learned Visual Tokenization and Perception Codecs

One class of related work seeks to replace raw pixels with a more compact or structured codeword. This includes, for example, VQ-VAE-style approaches [9] as well as BEiT-style tokenization schemes [10], in which the image is first converted into discrete tokens that are then used to train representations. However, in such systems the encoder itself is typically trained on large datasets and requires substantial compute, and the resulting representation remains part of a large ML pipeline rather than an independent perception layer.

TAPe takes a different stance. In this paper, TAPe is treated not as another learned tokenizer, but as a structured representation that defines relationships between perceptual elements prior to the main model. In this sense, TAPe is closer to a perception codec than to a standard learned embedding: the goal is not merely to compress the image, but to construct an input space in which much of the structure is already explicitly present.

## 3.2. Unified Backbones for Multiple Vision Tasks

A separate line of work aims to unify computer vision architectures, i.e., to use one backbone for multiple tasks – classification, detection, segmentation, or retrieval. In modern systems this role is most often played by large CNN or ViT-like models that are then fine-tuned or adapted to a specific task. In practice, however, such solutions typically remain task-specific at the level of the final architecture: different heads and training pipelines are used for detection, segmentation, and retrieval, and the final infrastructure still ends up split across a set of separate models.

TAPe+ML differs in that it unifies not only the backbone but the input space itself. The TAPe representation serves as a shared layer for multiple recognition tasks, with small specialized modules operating on top of it, coordinated within a single architecture. As a result, "unified backbone" here means not merely "the same network feeding different downstream heads," but a single perception space and a single recognition core for classification, detection, and segmentation.

## 3.3. Object Detection and Segmentation

Modern detectors from the YOLO and RF-DETR families are the closest to TAPe+ML in their application scope. These models set a strong practical baseline for COCO detection and serve as the standard point of comparison for new architectures. In particular, RF-DETR achieves a high level of quality on COCO and provides an important benchmark for comparing mAP50 and mAP50-95.

Nevertheless, such systems achieve their quality primarily through large backbones, millions of parameters, and intensive pixel-based training. TAPe+ML v3 takes a different path: moving from a pixel-tensor pipeline to a structured TAPe representation and a modular architecture with a submodel coordinator. In this sense, this work is not just another variant of the YOLO or DETR family, but offers an alternative line of development for detection, in which some of the complexity is shifted from model parameters to the level of representation and organization of perception.

## 3.4. Lightweight and Edge Vision Models

Another important area concerns lightweight models for edge deployment, such as MobileNet-like architectures [11]. Their main focus is reducing the number of parameters, cutting latency, and running on constrained hardware. However, most of these models remain specialized for a single task and do not address the fragmentation of the vision stack: even if the classifier and detector individually become more compact, this still does not yield a unified perception space or eliminate the need to maintain multiple pipelines.

TAPe+ML v3 is close to the edge-deployment direction in terms of its computational profile, but it differs in its underlying goal. Making the model compact was never a goal in itself. Compactness is a natural consequence of the model's structure – the only structure that could work with our data. The contribution of this work therefore lies not simply in model compression, but in a more general redesign of the vision pipeline around a structured representation.

## 3.5. Classical Visual Encodings

Before the dominance of deep learning, computer vision relied heavily on manual or semi-algorithmic image encodings: histograms, HOG [12], edge detectors such as Sobel and Canny [13], optical flow [14], and other descriptors. These methods are often fast or interpretable, but as a rule they do not provide a single representation equally suited to complex recognition tasks, and they do not scale well in quality to the level of modern SOTA systems.

In this paper, TAPe is loosely related to this class of methods in that it, too, is non-pixel and non-patch based, but it differs in purpose and level of generality. It is not another handcrafted descriptor: the TAPe representation serves as a shared layer across different vision tasks and then forms the basis for training a compact recognition model. A detailed comparison of TAPe with classical encodings and modern ML models on the task of segmenting video into scenes is given in Section 5.1.

## 3.6. Positioning of TAPe+ML

Existing work therefore typically covers only part of the property set that TAPe+ML claims. Learned tokenizers and self-supervised encoders provide useful insight but do not remove the dependence on large datasets and large models; YOLO- and RF-DETR-level detectors deliver strong quality but remain heavy and specialized for a single task; lightweight architectures reduce computational cost but do not form a unified perception space for multiple tasks. TAPe+ML occupies a distinct position in this landscape: it is a system in which structured perceptual representation, a unified recognition architecture, and practical data and compute efficiency are treated as parts of a single approach.

## 3.7. Summary of Related Approaches

The Table 1 summarizes the key differences between TAPe+ML and a few representative areas that this work intersects with.

*Table 1. Comparison of TAPe+ML with representative related approaches*

| **Method / family** | **Structured Perceptual Representation** | **One Backbone for Multiple Tasks** | **Strong Benchmarks** | **Compactness for Edge** |
|---|---|---|---|---|
| VQ-VAE / visual tokenizers | Partially: the image is encoded in discrete codes, but the tokenizer itself is trained as part of a large pipeline and remains inside it. | No: codes are usually used as part of specific pretraining / downstream pipelines, and not as a single recognition core. | Not the main goal: focus on improving views, not on a single compact system with SOTA for detection / classification. | No: Training and usage require significant resources and are not focused on ultra-small models. |
| BEiT / masked visual tokens | Partially: discrete visual tokens are used, but within a patch-based transformer trained on large datasets. | Partially: the pretrained backbone can adapt to different tasks, but downstream architectures remain specialized. | Yes, for representation learning and transfer, but not as a single, compact "all-in-one" recognition core. | No: a large transformer architecture, not an edge-first solution. |
| YOLO family | No: operates on pixel/tensor inputs, with features extracted inside the detector. | No: a strong detector, but not a single core for classification, segmentation, and retrieval without separate pipelines. | Yes: sets a strong practical detection baseline (COCO, etc.). | Partially: lightweight variants exist, but these remain detectors with millions of parameters. |
| RF-DETR | No: a transformer-based detector that operates on pixels, not on a perception codec. | No: focused on detection/segmentation, not a universal backbone for all tasks. | Yes: strong results on COCO, including AP50-95 > 60 for large variants. | No: substantially larger than ultra-compact edge models. |
| MobileNet / Light CNNs | No: pixel-optimized CNNs, without a separate perceptual layer. | No: models are lightweight but typically remain specialized for a single task. | Partially: competitive under tight resource constraints, but do not address the need for a single SOTA core across many tasks. | Yes: purpose-built for mobile and edge devices, with few parameters and low latency. |
| **TAPe+ML v3** | Yes: uses TAPe as a structured perceptual layer instead of raw pixels and standard patches. | Yes: the same TAPe space and recognition core are used for classification, detection, segmentation, and similarity search. | Yes: achieves mAP50 = 84.7% and mAP50-95 = 65.3% on COCO, plus strong classification results up to the ImageNet-1k level. | Yes: fewer than 100k parameters, designed to run on both regular and edge hardware. |

The external method families summarized in this table are represented by VQ-VAE [9], BEiT [10], YOLO [1], RF-DETR [2], and MobileNet [11].

## 3.8. Classical CV Encodings

Classical image representation methods include histograms, HOG, edge detectors, and optical flow. HOG describes gradient directions in the neighborhood of each pixel and works well for rigid objects such as pedestrians and faces, but struggles with deformable and complex scenes. Sobel/Canny detectors capture edges and carry mostly geometric information without explicit semantics; optical flow estimates inter-frame motion but is computationally expensive and sensitive to noise and artifacts. Simple brightness or color histograms are compact but fail to distinguish well between scenes with different objects and similar intensity distributions.

In this paper, TAPe is viewed as a step toward a more structured representation: rather than relying on local gradients or brightness distributions, it defines elements and relationships between them as the basis for recognition. For a detailed empirical comparison of TAPe against HOG, histograms, optical flow, and ML models on the task of segmenting video into scenes, see Section 5.1.

# 4. TAPe Representation and TAPe+ML Architecture

## 4.1. TAPe Representation and T-bits

Modern computer vision models based on CNNs and transformers operate on images represented as pixel tensors. Each pixel $I(x, y)$ stores only local brightness or color; the model must learn all structural relationships (edges, shapes, repeating patterns, object membership) on its own, relying on multi-layer convolutions or self-attention. This makes the representation both redundant and poorly structured: the same scene can have an enormous number of pixel realizations, and a significant portion of the computation is spent on recovering relationships between local values.

TAPe (Theory of Active Perception) takes a different approach. The image is described not by raw pixels, but by a set of TAPe filters (T-bits), each of which corresponds not to a single point but to a subset of the most informative, interconnected visual elements.

A T-bit is a subset of the most informative interconnected information elements. This definition matters because it immediately distinguishes a T-bit from a pixel. A pixel encodes only a local brightness or color value and carries no explicit information about its role in the scene. A T-bit, by contrast, is already a unit that includes relationships between elements, and thus carries more meaningful information for recognition. The TAPe representation is therefore organized from the outset as a structure of similarities and relationships, rather than a flat array of numbers.

Since the complete algorithm is proprietary to TAPe, the formal description in this paper can only be given at the level of input, output, and representational properties, not at the level of the full internal procedure for constructing T-bits.

Formally, the TAPe representation of an image $I$ can be described as a mapping

$$I \mapsto G(I) = (\{t_1, \dots, t_k\}, E)$$

where $t_i$ is a TAPe filter in a finite set of filters (T-bits) in a certain space $\Upsilon$, and *E* is the set of connections between them (a relational graph). Each T-bit $t_i$ corresponds not to a single coordinate $(x, y)$, but to a specific subset of pixels $R_i \subset \Omega$ and functions over them, selected according to TAPe rules. In practice $K \ll HWK$ is executed: the number of T-bits per image is orders of magnitude smaller than the number of pixels, and the TAPe index (the description $G(I)$) occupies orders of magnitude less memory than pixel- or gradient-based descriptions. This formulation does not claim to reveal the algorithm, but it establishes that the output of TAPe methods is not a flat feature vector or pixel tensor, but a **structured representation** consisting of elements and relationships.

The TAPe representation has several properties that fundamentally distinguish it from a pixel representation.

**Invariance to local transformations.**
The same set of T-bits describes an object under changes in lighting, minor deformation, and noise, whereas the pixel representation changes substantially. This is because the TAPe representation forms an inherently smooth space, i.e., the result is a smooth function. Put simply: the same object under different lighting conditions produces very similar descriptions, which lets us speak simultaneously of the stability of object descriptions while still being able to compute the lighting difference from that residual difference. In other words, there is no loss of essential information as such. As a result, the TAPe layer can act as a more stable "perceptual" foundation than raw pixels.

**Semantic stability.**
For two frames of the same scene, the TAPe indices (sets of T-bits and their relationships) are nearly identical even if the pixels differ due to camera motion or compression.In the video scene-segmentation experiment described in Section 5.1 TAPe was compared against classical encodings (histograms, HOG, optical flow) and modern ML models (including DINOv2) using the same HDBSCAN clustering algorithm [15] and identical parameters throughout. TAPe produced the most stable division of the video into scenes, with the lowest noise level, illustrating the ability of the TAPe representation to capture scene structure rather than just local signals.

**Conciseness and comparability.**
The TAPe index serves simultaneously as a compact representation and as a space for similarity search: the distance between TAPe indices directly corresponds to the semantic distance between scenes or images, without requiring additional training of large models on millions of examples. In the video experiment reported in Section 5.1, the TAPe index was built in 10-11 seconds and occupied less than 1 MB, while clustering took about 1 second. By comparison, classical methods (HOG, histograms, optical flow) and DINOv2 produced indices ranging from tens to thousands of megabytes and required hundreds to thousands of seconds to cluster the same video. These figures are not meant as a universal characterization for every task, but they illustrate the typical order-of-magnitude advantage of the TAPe representation in size and speed on video search tasks.

An important consequence is that the TAPe+ML architecture does not operate on raw pixels or patches in the usual sense, but on TAPe elements. The model receives an already structured, compact, and semantically rich representation as input, in which much of the work of "reconstructing structure" is performed at the level of TAPe methods rather than inside a deep network. In this sense, T-bits can be regarded as the elementary units of the TAPe language of perception: they replace the combination of raw pixels and heavy tensors, serving as the "atoms" of a unified recognition space in TAPe+ML.

## 4.2. TAPe+ML v3 Architecture

In the first version of TAPe+ML, the detection architecture was essentially monolithic: a single model, taking a TAPe representation as input, solved the task end-to-end and played the role of a "classic detector." In TAPe+ML v3, the architecture is fundamentally different: the system is built as a set of small, specialized models whose collaboration is managed by a submodel coordinator. Each submodel solves a simple, narrow problem, and the final detector emerges as the result of their coordinated interaction.

### 4.2.1. TAPe+ML v3 Submodels

**Background model**.
The first submodel is responsible for separating "uninteresting" areas (background, uniform textures, regions without objects) from potentially informative parts of the image. During development, we gradually moved away from processing the entire field of view toward working with a more compact set of TAPe patches and regions where objects can actually be located.

The background model takes a TAPe representation in the form of a Bag of Words and labels geometric segments as background/non-background, thereby reducing the load on the other submodels. Like all other submodels, it takes as input TAPe filters together with a description of their location. Using this data, the model successively builds several masks that are essentially 2D geometric primitives. These masks are merged into object contours, within which normals determine their orientation (i.e., what lies inside, and whether an "interior" exists at all). The final output is a description of the curves as splines.

**Pointer model.**
The second submodel localizes objects: it tells the coordinator exactly where candidate objects are located in the TAPe representation. Having abandoned transformers, TAPe+ML v3 uses local associations between TAPe patches together with a lighter object-search mechanism than global self-attention. The pointer model can be interpreted as a module that uses TAPe elements and the output of the background model to identify regions of interest (for example, centers or regions of future bounding boxes).

**Proto-cluster model (prototype / k-means).**
The third submodel is responsible for forming class prototypes. Instead of training a separate classifier on top of TAPe features via backpropagation, TAPe+ML v3 uses clustering (k-means) [16] to generate proto-descriptions of classes. TAPe elements or TAPe patches corresponding to objects of a given class are grouped into clusters, and the cluster centers (or their configurations) serve as class prototypes. At inference time, new objects are matched against these prototypes, and the class is determined by the structure of the TAPe space rather than by the parameters of a separate linear layer. This means the classifier is not trained as a separate parameterized module; it emerges from the geometry of the TAPe space and the results of k-means.

**Submodel coordinator.**
All submodels are managed by a coordinator, a module that controls the sequence and configuration of their invocations. The coordinator takes the TAPe representation as input and:

- passes it to the background model to filter out irrelevant areas;
- invokes the pointer model to localize objects in the remaining areas;
- uses the proto-cluster model to match objects with class prototypes;
- combines the results into final detection, classification, and (where applicable) segmentation predictions.

TAPe+ML v3 can automatically scale the number of submodels deployed for a given dataset: for more complex datasets, the coordinator can use more instances of individual submodels, while for simpler tasks it can rely on fewer modules. The configuration is also chosen based on loss behavior. When the losses begin to converge but the result is worse than required, the coordinator adds new submodels based on which losses show the worst convergence.

#### 4.2.2. Coordinator as an Alternative to a Monolithic Detector

The transition from a monolithic architecture to a system with a submodel coordinator matters for several reasons. First, it allows a complex detection and classification task to be split into a set of simple subtasks: background filtering, object localization, and prototype formation. Each module is simpler in structure and, as a result, can have fewer parameters. Second, the coordinator provides a foundation for adapting the architecture to a specific dataset: instead of changing the entire model, one can simply adjust the configuration and number of submodels.

This architecture is also consistent with the general idea behind TAPe: a significant part of the "intelligence" is shifted to the level of information representation and organization (the TAPe layer and submodel coordination), rather than residing solely in the parameters of one large neural network.

#### 4.2.3. Removing Gradient Descent from Classification

One of the most notable differences in TAPe+ML v3 is the replacement of classical gradient descent in the classification component with coordinate descent. Dependence on the learning-rate hyperparameter was itself a source of instability, so eliminating it required abandoning gradient descent in this part of the architecture. The switch to k-means-based classification on TAPe data produced an accuracy gain of about 3 percentage points in the described experiment.

Thus, in TAPe+ML v3:

- classification is performed via k-means in the TAPe space, matching against class prototypes;
- this component uses no backpropagation and requires no learning-rate selection;
- model calibration is shifted to the TAPe representation and prototype structure, rather than to optimizer settings.

This differs from standard modern architectures, where even with pre-extracted features, the final classification layer is almost always trained via gradient descent and depends on the learning rate and its schedule. In TAPe+ML v3, this source of uncertainty and potential instability is eliminated for the classification component.

#### 4.2.4. Architecture Parameters and Size

The total number of model parameters is under 100 thousand, with accuracy comparable to, and in places exceeding, RF-DETR 2XL on COCO. This applies to the complete coordinator-plus-submodels system, not just one of its parts. By comparison, RF-DETR 2XL has hundreds of millions of parameters, and many transformer-based detectors fall in the 20-100M parameter range or higher.

Thus, TAPe+ML v3 fits within a budget of fewer than 100k parameters; the architecture achieves this through the coordination of several small submodels rather than a single large detector. The smallest submodel is the proto-cluster model, and there are often several instances of it – on COCO, for example, three layers were deployed. The second-largest is the pointer submodel; on COCO, two instances were used, corresponding to different levels of refinement. The largest is the submodel separating object-like "things" from background. This background-separation submodel is the largest because most of the discrimination and description work happens there, given the specifics of our shared backbone design.

#### 4.2.5. How the TAPe+ML v3 Architecture Relates to the TAPe Representation

The TAPe+ML v3 architecture relies on the TAPe representation as a shared perception layer. All submodels – the background model, the pointer, the proto-clusters, and the coordinator – operate not on raw pixels or arbitrary patches, but on TAPe elements and their relationships. This means that:

- the coordinator operates on already structured "building blocks" with known relationships;
- the background and pointer models operate on the TAPe structure rather than on pixel tensors;
- class proto-clusters are built in the TAPe space rather than in the space of raw CNN or ViT features.

Together, the TAPe representation and the TAPe+ML v3 architecture form a system in which much of the complexity is shifted from the parameters of a large network to the level of representation and the composition of simple models.

## 4.3. Adaptation Modes

TAPe+ML supports several modes for adapting the model to a new task. In practice, three modes are used: head-only, backbone adaptation, and from-scratch. These modes differ in how much of the system is changed for the new task and how strongly the model adapts to the new data domain.

### 4.3.1. Head-Only

In head-only mode, the base representation and backbone remain frozen, and only the top application layer (the "head") responsible for a specific task – such as classification or detection – is trained on the new task. This mode is naturally suited to cases where the new task is close to the domain on which the backbone was originally tuned: objects are visually well defined, capture conditions are standard, and angles and lighting do not differ substantially from the base training distribution.

In such cases, we can expect the existing TAPe representation and the already-formed backbone to contain sufficient information, requiring only a light adjustment of the output layer. In this sense, head-only is the cheapest and fastest adaptation mode: it requires the fewest trainable parameters and minimal computational cost.

However, head-only is sensitive to domain shift. If the backbone was formed on one type of data while the new task belongs to a different visual domain, adapting the head alone may not be sufficient. In an industrial pilot involving detection of contaminated ore, head-only mode produced substantially lower accuracy than backbone adaptation; detailed quantitative results are given in Section 5.4. This result is an important indicator that even a strong TAPe representation at the upper layer is not enough if the backbone itself remains biased toward the original domain.

### 4.3.2. Backbone Adaptation

In backbone adaptation mode, both the backbone and the application head are reconfigured for the new task. This mode is used when the task belongs to an out-of-distribution (OOD) domain, i.e., one that differs substantially from the data on which the original representation was formed. For example, in the industrial pilot, the standard backbone – biased toward COCO-type objects – transfers poorly to images from a production environment, where the visual structure of objects, lighting, textures, and patterns differ from typical everyday scenes.

This is where what could be called COCO bias arises in practice: if the backbone was formed on the visual world of COCO, it is by definition better aligned with people, vehicles, animals, road scenes, and other standard classes than with industrial objects such as contaminated ore. Under these conditions, head-only applies an output layer on top of a representation that is itself already biased, and therefore performs poorly. Backbone adaptation removes this limitation by allowing the internal representation itself to be rebuilt for the new domain.

In the industrial pilot, this mode delivered a noticeable accuracy gain over head-only; detailed quantitative results for all modes are given in Section 5.4. This shows that adapting the backbone is not merely an additional refinement, but a necessary step for OOD tasks.

#### 4.3.3. From Scratch

The third mode is from scratch, i.e., training the model entirely from zero for the new domain. Here, the original backbone is not assumed to already contain useful information for the target task; the entire architecture is trained anew. This mode is the most costly in time and compute, but in principle offers maximum flexibility, since it inherits no bias or limitation from the original domain.

This mode is needed when the new domain is fundamentally unlike the previous one and partial adaptation proves insufficient. This can be relevant for entirely new sensing environments, non-standard image types, or domains where the original representation barely transfers at all.

#### 4.3.4. Why COCO Bias Arises

The choice of training mode is directly tied to how strongly the original backbone is biased toward the source dataset. If the backbone was formed on a visual world close to COCO, it already contains useful priors for tasks similar to standard categories (people, vehicles, animals, everyday objects).

In an industrial task, however, the visual patterns can be entirely different. Contaminated ore, surface defects, production-line elements, or specific materials need not resemble COCO objects in structure, texture, or illumination distribution. In this case, the backbone ends up biased toward a "foreign" distribution, and adapting only the head is not enough. This is precisely why backbone adaptation becomes a necessary mode in OOD scenarios, with from-scratch training as the extreme case of a full system rebuild.

#### 4.3.5. Auto-Annotate

Separately, the auto-annotate mode – i.e., semi-automatic labeling – plays an important role in TAPe+ML. After a certain number of images (for example, 10-15) are annotated, the user can feed them to a simplified version of the TAPe+ML v3 model, which, after quickly training on this small labeled set, marks up possible detections above a certain (low) confidence threshold on the remaining images. The user can either confirm these detections or edit them, enabling an iterative workflow. At each round of auto-labeling, the user sees which objects the model is less confident about and annotates more images containing that object, yielding better detections for downstream images.

From a practical standpoint, auto-markup solves two problems at once. First, it lowers the barrier to entry: instead of full manual annotation, a small labeled subset suffices. Second, it accelerates the transition from pilot to working system, since the initial model begins expanding the annotation itself, and the human role is reduced to reviewing and correcting the automatically found objects.

### 4.3.6. Summary of Modes

Thus, the three modes form a natural scale for adapting TAPe+ML to a new task. Head-only suits tasks close to the original domain and minimizes fine-tuning cost. Backbone adaptation is needed for OOD scenarios and allows the representation itself to be rebuilt for the new visual environment. From-scratch is the extreme mode for fundamentally new domains where even an adapted backbone may prove insufficient. Auto-markup complements these modes by lowering annotation cost and speeding up transfer to applied tasks.

As part of our experimental study, we also performed a comparative analysis on the standard RF100-VL [17] dataset Lacrosse-object-detection-uxkt [18], using the uavdet-small dataset [19] from the same collection as an OOD scenario. For the three model configurations (head, backbone, and TAPe+ML v3), the resulting mAP values were 64.1, 68.2, and 72.4, respectively, in the standard setting, and 45.6, 52.7, and 58.5 in the OOD setting.

These results show that, in the standard setting, absolute quality values start higher, and the relative gain from head to TAPe+ML v3 is about 8 percentage points, whereas for the OOD scenario the gain reaches almost 13 percentage points, with a substantial gap between the head and backbone levels. This indicates that the proposed TAPe+ML v3 delivers the largest benefit precisely under distribution shift, where simply increasing architectural complexity (moving from head to backbone) is less effective.

*Table 2. RF100-VL experiment results on the Lacrosse-object-detection-uxkt dataset and the out-of-distribution uavdet-small scenario*

| Configuration | Dataset | mAP |
|---|---|---|
| Head | Standard (Lacrosse) | 64.1 |
| Head | OOD (uavdet-small) | 45.6 |
| Backbone | Standard (Lacrosse) | 68.2 |
| Backbone | OOD (uavdet-small) | 52.7 |
| **TAPe+ML v3** | **Standard (Lacrosse)** | **72.4** |
| **TAPe+ML v3** | **OOD (uavdet-small)** | **58.5** |

# 5. Experiments

This section describes experiments in which TAPe and TAPe+ML v3 were compared against classical methods and modern computer vision models. TAPe serves as the representation layer (TAPe indices, T-bits), and TAPe+ML v3 serves as the recognition architecture built on top of this representation.

## 5.1. Video Scene Detection (Interstellar)

The video segmentation experiment into meaningful scenes was the first public test demonstrating the benefits of TAPe as a representation. The task was to split a fragment of the film Interstellar into scenes using the same clustering algorithm (HDBSCAN) across different video representation methods. The same video, the same clusterer, and the same parameters were used for all methods, so that any differences would be attributable solely to the quality and compactness of the representation.

The experiment compared more than 13 machine learning models (ConvNeXt [20], SwinT [21], ViT, EfficientNet [22], DINOv2, MobileNet, and others) and 6 classical methods (HOG, Sobel, Canny, optical flow, histograms, and Fourier-based methods). TAPe served as a separate method that converts video into a TAPe index – a compact representation consisting of T-bits and their connections, without using neural networks to build features.

Hardware used in the experiments:

CPU: Intel Xeon E5-2697 v2 @ 2.70 GHz

RAM: 256 GB, DDR3, 1600 MT/s

The summary results (indexing, clustering, file size, cluster quality) can be described as follows:

- TAPe: video indexing takes 10-11 seconds, clustering takes about 1 second, and the TAPe index size is under 1 MB; TAPe also produced the most stable division of video into scenes (frames from the same scene fall into a single cluster, with minimal "noisy" clusters).
- DINOv2 ViT-B/14: indexing takes about 5220 seconds on 16 CPU cores, clustering about 21.5 seconds, index size about 9.7 MB; cluster quality is good but worse than TAPe on scenes lacking pronounced objects.
- HOG: indexing takes about 530 seconds, clustering about 2226 seconds, file size about 1.7 GB; cluster quality is unstable, with scenes often split or merged incorrectly.

- Optical flow: indexing takes about 2959 seconds, clustering about 8739 seconds, file size up to 3.2 GB; scene segmentation quality is low, especially on static and slow-moving footage.
- Histograms (e.g., 8 levels): indexing takes about 27 seconds, clustering about 13 seconds, index size about 6.5 MB; quality is poor, with different scenes sharing similar brightness distributions falling into the same cluster.

The DINOv2 result is particularly telling: the model performs well on scenes with pronounced objects but struggles with "empty" or dark scenes (space, dark transitions), where the classification-oriented CLS token fails to provide a stable structure for clustering. TAPe, by contrast, relies on scene structure and relationships between T-bits, and therefore separates even such episodes more effectively.

Video clustering results (scene splits) cannot be included in the article format, but a comparative freeze-frame grid (e.g., a 2x2 grid: TAPe, DINOv2, HOG, histograms) can illustrate the difference, with annotations showing which frames fell into the same cluster for each method. The experiment description should also explicitly state the hardware on which timing measurements were performed (CPU type, number of cores, presence or absence of a GPU), since this affects the interpretation of indexing and clustering speed.

## 5.2. Object Detection on COCO

In this section, TAPe+ML v3 is evaluated on the standardized COCO benchmark and compared against strong RF-DETR- and YOLO-level detectors. The goal is to show that a detector built on a TAPe representation with fewer than 100 thousand parameters can achieve accuracy comparable to modern models built on transformers and large CNNs.

### 5.2.1. Experimental Setup

The standard COCO set (train2017 / val2017) with 80 object classes is used to evaluate detection. TAPe+ML v3 takes a TAPe representation of images as input rather than raw pixels; this representation is described in Section 4.1. The TAPe+ML v3 architecture includes a submodel coordinator and several small specialized modules (background, pointer, proto-clustering), with a total parameter count remaining below 100k.

The evaluation metrics are the same as those used for RF-DETR and YOLO: mAP50 and mAP50-95 (COCO AP).

- mAP50: average detection accuracy at an IoU threshold ≥ 0.5 (an object is considered found if the predicted box overlaps ground truth by at least 50%).
- mAP50-95: average accuracy across 10 IoU thresholds from 0.5 to 0.95 in increments of 0.05 – a stricter metric sensitive to localization accuracy and the structure of predicted boxes.

In addition, oracle classification accuracy is measured: classification accuracy assuming the bounding boxes are set perfectly (oracle boxes). This metric reflects the quality of TAPe embeddings and the classification component independent of localization errors.

### 5.2.2. Results for TAPe+ML v3

The final results for the TAPe detector on COCO can be summarized as follows:

- TAPe+ML v3:
    - mAP50 = 84.7%;
    - mAP50-95 = 65.3%;
    - total number of parameters < 100k;
    - latency is reported separately below under the stated GTX 1070 Ti measurement protocol.

For comparison, RF-DETR 2XL (one of the strongest publicly available transformer detectors) [2] achieves an mAP50-95 of about 60.1 on COCO, with roughly 127 million parameters and a latency of about 17 ms per image on comparable hardware. Modern YOLO variants (e.g., YOLOv9 [23]/v10 [24]) typically fall in the 40.1-56.9 mAP50-95 range, depending on the variant (nano/small/medium) and the number of parameters in millions. Notably, YOLO26n is generally regarded as a slight regression compared with earlier nano models in the series; the YOLO11n model, for instance, achieves higher results, around 41.

TAPe+ML v3 in this context:

- shows mAP50 above RF-DETR 2XL and strong YOLO models;
- exceeds RF-DETR 2XL on mAP50-95 (65.3 vs. ≈60.1);
- while using fewer than 100k parameters – 2-3 orders of magnitude fewer than RF-DETR 2XL and many YOLO variants.

Oracle classification accuracy for TAPe+ML on COCO reaches about 87.3%: given perfect boxes, TAPe embeddings and the classification component correctly identify the class in about 87.3% of cases. This indicates that the TAPe embeddings themselves are highly informative even before accounting for localization errors. A substantial part of the remaining headroom lies not so much in the classification component as in more accurate box localization and post-processing optimization.

*Table 3. COCO object-detection comparison of TAPe+ML v3 and public baselines*

| Model | COCO mAP50 | COCO mAP50-95 | Parameters | Delay (ms, bs=1) | Note |
|---|---|---|---|---|---|
| **TAPe+ML v3** | **84.7** | **65.3** | **< 0.1 M** | **≈10-12 (GPU / CPU)** | **TAPe Detector, third version** |
| RF-DETR-2XL [2] | 78.5 | 60.1 | 126.9 M | 17.2 (T4) | Transformer det., Roboflow |
| RF-DETR-M [2] | 73.6 | 54.7 | 33.7 M | 4.4 (T4) | Medium size RF-DETR |
| YOLO11-M [1] | 64.1 | 48.6 | 20.1 M | 5.1 (GPU) | Representative of the YOLO11 family |

It is important to keep in mind that the latency figures for RF-DETR and YOLO are taken from their official benchmark tables, where measurements are performed under standardized conditions (e.g., NVIDIA T4, TensorRT, batch size 1).

**Model Latency**

Latency was measured on the same machine with an NVIDIA GTX 1070Ti 8GB GPU; TAPe+ML v3 shows a latency of 10.8 ms per frame (including TAPe segmentation), which is faster than the compact YOLO26s configuration (12.3 ms) and substantially faster than RF-DETR-Small (41.0 ms).

This measurement refers to the detection-stage benchmark; the full end-to-end pipeline, including detection, classification, final instance segmentation, and post-processing, is reported separately in Section 5.4.7.

*Table 4. Detection-stage inference latency on NVIDIA GTX 1070 Ti at batch size 1*

| **Model** | **Time per frame, ms** | **Hardware Platform** |
|---|---|---|
| **TAPe + ML v3** | **10.8** | **NVIDIA GTX 1070Ti 8Gb (GPU)** |
| YOLO26s [1] | 12.3 | NVIDIA GTX 1070Ti 8Gb (GPU) |
| RF-DETR-Small [2] | 41.0 | NVIDIA GTX 1070Ti 8Gb (GPU) |

### 5.2.3. Learning Dynamics and Architectural Changes

The path to the final TAPe+ML v3 involved several stages, reflecting the gradual restructuring of the detection architecture:

- use of centroid-based schemes and simple models with ~115k parameters;
- addition of high-quality TAPe attributes to the COCO subset;
- training TAPe embeddings in a self-supervised style (iBot-like approaches [25]);
- abandoning transformers in favor of local associations of TAPe patches and a simplified, lighter detector;
- adding boundary-based segmentation and gradient-free k-means classification in the TAPe space.

In the course of this evolution, the first submodel was completely redesigned: we moved away from a simple "background vs. non-background" box classification toward a more sophisticated search for correct contours, along with better initial object localization, reducing the number of catastrophic failure cases, such as False Negatives, in this submodel – which would otherwise propagate into downstream errors – to zero.

In the final v3 version, the architecture combined these elements into a system with a submodel coordinator and a total parameter count under 100k, while improving mAP50-95 relative to earlier TAPe+ML versions and RF-DETR 2XL.

For timing measurements, we used a GTX 1070 Ti 8GB GPU, batch size = 1, at the original resolution but with a 1024-pixel long-side cap, using the standard pycocotools framework [26] for computing results.

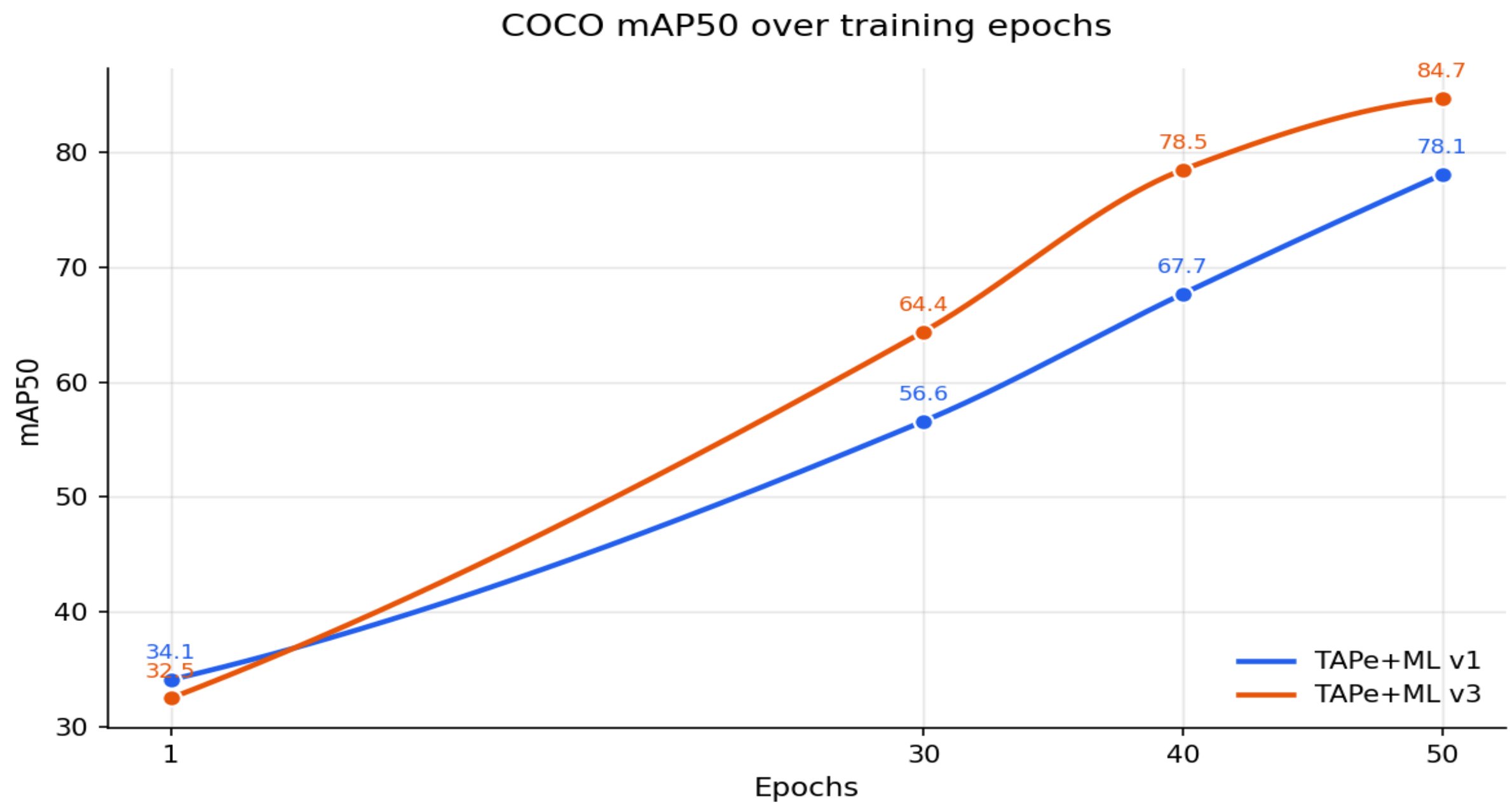

*Figure 4. Learning dynamics and architectural evolution of TAPe+ML v3 on COCO*

## 5.3. Classification

In this section, the TAPe representation and TAPe+ML architecture are evaluated on classification tasks at varying levels of difficulty: from compact datasets (Imagenette, MNIST) to large-scale ImageNet-1k. The goal is to show that TAPe's benefits extend beyond detection to classical recognition tasks, and that the same architecture can operate at different scales with an extremely small parameter budget.

### 5.3.1. Imagenette

Imagenette is a subset of ImageNet with 10 classes, commonly used for quick experiments and architecture comparisons. In the experiments described here, a 3-layer CNN with roughly 516 thousand parameters was trained on 10% of the sample without augmentation. Two configurations were compared:

- input is the TAPe representation (TAPe features built on top of the original images);
- input is raw pixels (baseline), with an identical architecture and training setup.

Results:

- the TAPe variant achieves about 92% validation accuracy;
- the raw-pixel baseline achieves about 47% under the same conditions.

This experiment is particularly telling, because:

1. the architecture and number of parameters are identical;
2. the amount of data and the training regime are identical;
3. the only difference is the representation of the input data.

Thus, the near two-fold difference in accuracy (92% vs. 47%), all else being equal, makes the contribution of the TAPe representation to classification quality strikingly clear.

### 5.3.2. MNIST

MNIST [27] is treated as a simple but well-studied benchmark for evaluating architecture convergence with a limited amount of data. In the TAPe+ML experiments, a stricter-than-usual split was used: 40% of the data for training and 60% for validation (as opposed to the typical 60/40 or 80/20).

Under these conditions, TAPe+ML achieved about 98.5% accuracy by the 10th epoch, with a smooth learning curve free of significant fluctuations despite the reduced training set. This shows that the architecture built on top of the TAPe representation maintains stable convergence even with a less favorable train/val ratio, and that TAPe features enable efficient use of a limited number of examples.

### 5.3.3. ImageNet-1k Top-1

To assess the scalability of TAPe+ML to large datasets, we evaluated the classification task on ImageNet-1k.

Modern large models are used as reference points:

- DINOv2 ViT-g/14: Top-1 ≈ 86.92%, about 1.1B parameters;
- ConvNeXt V2-Large [28] (IN21k → 1k): Top-1 ≈ 87.26%, about 198M parameters;
- DINOv3 ViT-7B/16 [29]: Top-1 ≈ 88.4%, about 7B parameters.

Results for TAPe+ML v3 (fewer than 100 thousand parameters):

*Table 5. Top-1 accuracy of TAPe+ML v3 on ImageNet-1k and distribution-shift evaluation sets*

| **Dataset** | **TAPe+ML v3 Top-1 (%)** |
|---|---|
| ObjectNet | 78.6 |
| ImageNet v2 | 80.2 |
| ImageNet ReaL | 89.9 |
| ImageNet-R | 90.3 |
| ImageNet-Sketch | 75.8 |
| ImageNet-1k | 88,1 |

The training and evaluation protocol fully follows the standard ImageNet-1k train/validation split and image preprocessing. Training was performed from scratch.

### 5.3.4. Comparison Tables: TAPe+ML v3 vs. SOTA Models

Recognition quality of TAPe+ML v3 against four strong baselines (ResNet-50 [30], ConvNeXt-B, DINOv2, DINOv3) on five standard distribution-shift benchmarks: ObjectNet [31], ImageNet-V2 [32], ImageNet-ReaL[33] , ImageNet-R [34], and ImageNet-Sketch [35]. Values are Top-1 accuracy; TAPe+ML v3 shows performance comparable to the DINOv2/v3 foundation models and superior to classical CNN/ConvNet backbones on challenging OOD sets, while remaining orders of magnitude more compact in parameter count.

*Table 6. Top-1 accuracy on distribution-shift benchmarks: TAPe+ML v3 and reference models*

| Dataset | **TAPe+ML v3 (%)** | ResNet-50 (%) | ConvNeXt-B (%) | DINOv2 ViT-B/14 (%) | DINOv3 ViT-7B/16 (%) |
|---|---|---|---|---|---|
| ObjectNet | **78.6** | 38.1 | 42.0 | 72.0 | 79.0 |
| ImageNet v2 | **80.2** | 76.3 | 75.0 | 79.0 | 81.4 |
| ImageNet ReaL | **89.9** | 91.1 | 89.0 | 86.0 | 90.4 |
| ImageNet -R | **90.3** | 81.3 | 69.0 | 78.8 | 91.1 |
| ImageNet -Sketch | **75.8** | 30.9 | 55.0 | 62.5 | 71.3 |

Number of parameters (in millions) and approximate floating-point operations (GFLOPs) per 224×224 image. TAPe+ML v3 achieves DINO-family-level OOD robustness at a budget of under 0.1M parameters and about 0.1 GFLOPs – several orders of magnitude more efficient in "robustness per parameter / per FLOP" than large CNN and ViT foundation models.

*Table 7. Parameter count and approximate computation per 224 × 224 image*

| Model | Params M | FLOPs_G_224sq |
|---|---|---|
| **TAPe+ML v3** | **0.1** | **0.1** |
| ResNet-50 | 25.6 | 4.1 |
| ConvNeXt-B | 89.0 | 15.4 |
| DINOv2 ViT-B/14 | 85.0 | 50.0 |
| DINOv3 ViT-B/16 | 86.0 | 12.5 |

## 5.4. Segmentation

### 5.4.1. How Segmentation Works Inside TAPe+ML v3

In our architecture, segmentation serves as a separate background-search submodel. It draws closed object boundaries at the pixel level, and this region becomes what the other submodels "look at" for localization and classification - i.e what they consider to be foreground.

**The segmentation submodel operates in three stages:**

First, all potential image boundaries are marked on the backbone features using a trained contextual mechanism that recovers even weak or blurry edges when they are contextually meaningful (for example, the model does not treat an object's shadow as a boundary, but does recognize the faint edge of an object against a similarly colored background).

These boundaries – essentially polylines – are then merged into polygons. If a set of lines does not close into a contour, it is discarded; the model leaves no "dangling" or unclosed fragments.

In the final stage, the interior surfaces of the contours are merged so that the number of segments per object is minimized: eyes become part of the head, buttons become part of the jacket, rather than remaining separate segments. The resulting contours are already topologically correct – free of random holes or gaps – because masks are not defined parametrically but constructed via Boolean operations on the generated regions. Holes in masks appear only where the object genuinely has them (for example, the handle of a pot or the gap in a chair).

Technically, this is implemented as a single segmentation head whose intermediate layers have their own auxiliary heads during training (for separately optimizing each of the three stages); at inference time, all of this is discarded, leaving only the final contour output. Importantly, masks are constructed without being tied to a specific object class. This submodel's task is to find meaningful closed contours in general, rather than learning a mask shape for each class separately, as is often done in detectors with a fixed category set.

### 5.4.2. A Shared Training Objective for Detection and Segmentation

Segmentation is tightly coupled with the rest of the TAPe+ML submodels: the entire system shares a single loss with several components, each scored for its own task but optimized jointly. This means detection training automatically trains segmentation, and vice versa, without an additional training pipeline or additional data beyond what detection already requires.

This coupling works both ways, and quite concretely rather than abstractly. Detection benefits from additional guidance from segmentation: if a mask is slightly wider than the object's true boundaries (by a couple of pixels), this is an almost negligible loss for the mask metric (a fraction of a percent of IoU), whereas for a bounding box the same error can cost a significant percentage of accuracy – so segmentation gives detection a more precise object geometry than it could obtain directly.

The reverse effect involves merging disparate segments. If a person in a photo is standing behind railings, segmentation sees several separate regions divided by the railing, and detection uses this information to recognize that it is still a single object and draws one common bounding box.

### 5.4.3. Why We Chose LVIS Over COCO

For segmentation, we needed pixel-perfect masks, so we used LVIS as the primary dataset for training masks. LVIS masks are annotated with high precision, without rough approximations or overlapping polygons. In COCO instance segmentation, masks have historically been defined as simplified polygons that do not always perfectly enclose an object's contour and sometimes physically intersect incorrectly between different objects – in other words, the dataset itself is poorly suited as a reference for training accurate contour-level segmentation.

However, LVIS has a distinctive feature: it is a federated dataset – not every object in every image is annotated, only some of them. Because of this, one cannot simply treat "everything unannotated" as background, since there may be real objects that were simply not annotated. LVIS therefore gives us only positive masks (where an object boundary is definitely present), and for negative examples (where boundaries definitely should not be present) we used panoptic COCO [36], in which the entire image is divided into segments completely and without gaps, allowing us to safely select categories where objects of our type clearly cannot appear. This mixed strategy is only needed during backbone pre-training; afterward, for further training on new datasets, these manipulations no longer need to be repeated – the model remains dependent on the quality of LVIS masks, but not on the LVIS dataset itself.

To avoid contaminating our metrics, we used the same set of images for LVIS and COCO – the images are the same, only the masks differ. In total, we used 5 thousand images for training (a little under 5% of the dataset), which was sufficient for boundary formation.

### 5.4.4. LVIS: 97-98% Object Coverage and Mask Accuracy at AP@75

Our LVIS result is 82.5% AP under the LVIS methodology, calculated on a combined val + minival split. We included minival in validation to avoid using it in training, since minival images are valid COCO images and training on them would contaminate the COCO results.

A more intuitive and, in our view, more honest metric than AP itself is how many objects receive a meaningful mask at all, and with what accuracy. We compared ourselves against YOLO without additional training on LVIS (i.e., on equal, zero-shot terms with respect to this dataset), on the same val split (~25 thousand images), using the same Mask IoU / AP@75 metric (the proportion of objects whose predicted mask falls within the true object's boundaries with IoU ≥ 0.75).

*Table 8. LVIS mask coverage and mask accuracy at IoU 0.75*

| **Metric** | **YOLO metric (without LVIS training)** | **TAPe + ML (before additional training)** | **TAPe+ML (after additional training)** |
|---|---|---|---|
| Covering objects with masks | ≈44% | ≈97% | **≈98.2%**** |
| Percentage of masks with AP@75 (IoU ≥ 0.75) | ≈21% | ≈72% | **≈74%** |
| Average Mask IoU (if "forced" to build a mask) | ≈0.135 | 0.345 | **0.856*** |

** Slightly higher than the final values, because classification is not included in the IoU.*

** The improvement from additional training is modest precisely because the starting values are already high – the main gain is on small objects, where an error of even one pixel along the contour immediately translates into tens of percent of IoU loss (simply because a small object has few pixels to begin with).

This difference in coverage is primarily architectural. YOLO-Seg constructs the mask not as an independent entity, but as a combination of a fixed set of trained prototypes (typically 32) at roughly a quarter of the original image resolution – for example, 160×160 for a 640×640 input. This coarse proto-mask is then upsampled to the final resolution. Combined with feature-map strides (typically 2-4x downsampling at different levels), the resulting mask boundary accuracy for such models falls in the range of roughly 8-16 pixels – the contour in this sense is not "located" but reconstructed by interpolating a fairly coarse grid. Our model works differently: the contour is built directly at the pixel level through the three stages described above, rather than by stretching prototypes, which yields fundamentally different coverage and accuracy.

### 5.4.5. TAPe+ML Results on COCO and RF100-VL

The segmentation-enabled TAPe+ML v3 system achieves box-detection mAP50 of 85.2% and box-detection mAP50-95 of 65.3% on COCO; it achieves box-detection mAP50-95 of 68.3% on RF100-VL.

For context, the strongest large model in the RF-DETR family, RF-DETR-2XL, achieves 60.1% mAP50-95 on COCO and 63.2% mAP50-95 on RF100-VL with a substantially larger number of parameters (126.9M vs. ~0.1M for us).

*Table 9. Box-detection mAP50-95 on COCO and RF100-VL*

| Model | COCO mAP50-95 | RF100-VL mAP50-95 | Parameters |
|---|---|---|---|
| **TAPeML v3** | **65.3%** | **68.3%** | **~0.1M** |
| RF-DETR-2XL | 60.1% | 63.2% | 126.9M |
| YOLO26-X | 56.9% | 60.0% | 56.9M |

It should be emphasized that segmentation comparisons require care: our contour coverage and accuracy figures (97-98% / 72-74%) were computed using the LVIS mask-coverage / AP@75 methodology, not the standard COCO instance-segmentation Mask AP used, for example, for RF-DETR-Seg (49.9% Mask AP on 2XL) or for large, non-real-time models such as Mask DINO Swin-L [37] (54.5% Mask AP, COCO val) and Co-DETR [38] (56.6% Mask AP, COCO val). These are different metrics measuring different things, and a direct comparison of "our % vs. their Mask AP" would not be fair.

### 5.4.6. COCO Instance Segmentation: RF-DETR, YOLO26, and TAPe+ML v3

For COCO instance segmentation, we compare against strong real-time models specifically positioned as SOTA for masks: RF-DETR-Seg 2XL and YOLO26-X-Seg.

Using the same standard COCO metrics (Mask AP50 and Mask AP50-95), our results are as follows:

*Table 10. COCO instance-segmentation results measured by Mask mAP50 and Mask mAP50-95*

| Model | Mask mAP50 | Mask mAP50-95 |
|---|---|---|
| RF-DETR-Seg-2XL [2] | 73.1 | 49.9 |
| YOLO26-X-Seg [1] | 71.6 | 46.8 |
| **TAPe+ML (instance seg)** | **80.7** | **58.4** |

This table shows that our single TAPe+ML model with built-in segmentation delivers a noticeably higher Mask AP at both AP50 and AP50-95 than the specialized, larger RF-DETR-Seg-2XL and YOLO26-X-Seg models, while remaining an order of magnitude lighter in parameters and compute.

#### 5.4.7. Speed: 12 ms for the Entire Pipeline

A distinguishing feature of our segmentation approach is that it does not trade accuracy for speed. The average speed of the entire pipeline (from input image to final masks and boxes, including backbone, detection, classification, segmentation, and post-processing) is about 12 ms. Measurements were taken on images at the original COCO resolution, with a 1024-pixel long-side cap (to prevent panoramic images from skewing the statistics), on a GTX 1070Ti 8GB GPU. On a typical CPU, speed remains within the same order of magnitude, since the computations are parallelized similarly.

As the number of objects in a scene increases, speed decreases, but only slightly, because all segmentation operations run within the same GPU kernel, with comparable parallelization available on CPU. This makes the model suitable for edge and online scenarios that require both contour accuracy and real-time responsiveness.

## 5.5. Industrial Pilot: OOD Detection and Classification in a Production Setting

Beyond standard benchmarks, TAPe+ML was tested in an industrial pilot on detecting ore blockages from conveyor-belt images. This experiment matters because it evaluates the model not on data close to COCO or ImageNet, but in an out-of-distribution (OOD) scenario, where the visual structure of objects, lighting, background, and class distribution differ substantially from standard academic datasets. Under such conditions, it becomes especially clear how well the model transfers to a new domain and whether it can operate with extremely small amounts of data.

### 5.5.1. Problem Statement

The pilot addressed detecting and classifying ore blockages in conveyor-belt images. This is a two-class industrial problem in which the objects do not resemble typical COCO categories – they are not people, vehicles, animals, or household items, but belong to a specific visual domain of the production environment. Additional complexity arises from non-standard camera angles, variable lighting, non-uniform material texture, and a limited amount of annotated data.

This is precisely why the experiment is considered an OOD scenario. The model, initially biased toward COCO-like objects, must adapt to a new visual world in which the important features do not coincide with the usual general-purpose categories. This makes the pilot especially informative for assessing how well TAPe+ML suits real edge deployment in production settings.

### 5.5.2. Compared Modes

The pilot compared several training modes and baseline approaches:

- YOLO 26s baseline [1] – a standard detection model serving as an external reference point;
- TAPe+ML head-only – only the applied head adapts; the backbone remains frozen;

- TAPe+ML + backbone adaptation – both the head and the backbone adapt to the new task;
- TAPe+ML + backbone adaptation + NMS [39] – additional post-processing to improve final prediction quality.

This comparison makes it possible to separately evaluate the impact of the TAPe representation itself, the adaptation mode, and the final post-processing stage.

### 5.5.3. Main Results

The pilot results can be summarized as follows:

*Table 11. Industrial OOD pilot results by adaptation mode and training-data volume*

| Method | Data volume | Accuracy* |
|---|---|---|
| YOLO 26s Data Volume (baseline)method | 30 images | ≈28% |
| TAPe+ML, head-only | 30 images | ≈45% |
| TAPe+ML, backbone adaptation | 30 images | ≈65% |
| TAPe+ML, backbone adaptation | 85 images | ≈85% |
| TAPe+ML, backbone adaptation + NMS | 500 images | 85.7-96% |

**The "Accuracy" metric above stands as a binary detection designation of detections above IoU >= 0.5 with the correct Top-1 classification prediction.*

For the final configuration with 500 images and NMS, results are also broken down by class: about 100% for one class and about 89.97-90% for the other. These values are already close to the level required for practical use in industrial quality-control systems.

### 5.5.4. Interpretation of Results

First, the YOLO 26s baseline, which is tuned toward COCO, shows low accuracy (≈28%) with a limited amount of data. This is expected: a YOLO backbone trained on standard general-purpose datasets does not transfer well to the visual domain of contaminated ore. In this sense, the experiment illustrates the COCO-bias effect: a strong general-purpose model can lose quality dramatically once the target objects and capture conditions fall outside the distribution on which it was formed.

Second, TAPe+ML in head-only mode raises accuracy to ≈45% on the same 30 images. Even without adapting the backbone, this is significantly better than YOLO 26s, demonstrating the advantage of the TAPe representation with a small number of examples. However, this result is still insufficient for practical deployment, indicating that adapting the top layer alone is not enough if the internal representation is not adapted to the new domain.

Third, enabling backbone adaptation delivers the most important gain: ≈65% accuracy on the same 30 images, rising to ≈85% when the sample is expanded to 85 images. Thus, TAPe+ML with an adapted backbone is already more than twice as accurate as the YOLO baseline on 30 images (≈65% vs. ≈28%). This is a key practical result of the pilot: TAPe+ML's advantage stems not only from being small or theoretically "interesting," but from genuinely transferring better to a new industrial domain.

Fourth, increasing the training sample to 500 images and applying NMS brings accuracy into the 85.7-96% range. This shows that the architecture not only outperforms the baseline at the very-low-data stage, but also continues to improve steadily as data accumulates, approaching a level acceptable for industrial operation.

### 5.5.5. Why TAPe+ML Outperforms YOLO in the OOD Scenario

The outcome of the industrial pilot can be attributed to two factors. The first is the structured TAPe representation, which from the outset works not with raw pixels but with more meaningful elements of perception. The second is the ability to adapt the backbone to a specific production environment, rather than merely reconfiguring the final head.

For YOLO 26s, the main limitation in this pilot relates specifically to transfer: its architecture and backbone are optimized for a different image distribution, in which objects and scenes are statistically closer to COCO. In the industrial ore problem, this inductive bias is a limitation rather than an advantage. TAPe+ML, in contrast, benefits from a combination of compact architecture, structured representation, and the ability to rapidly adapt the backbone to a new domain, even with only dozens of images.

Thus, TAPe+ML transfers to OOD tasks substantially better than the standard COCO-oriented baseline, while retaining high data efficiency in a real production environment. In our view, this makes the experiment one of the strongest arguments for the practical value of TAPe+ML.

#### 5.5.5.1. The training protocol for backbone adaptation is as follows:

**Number of epochs**: 15

**Batch size**: 64

**Early stopping**: no improvement for 3 consecutive epochs after epoch 5 triggers a training stop.

**Training strategy**: OneCycleLR [40], with the peak learning rate occurring at epoch 5 (i.e., at 34% of total epochs), followed by a cosine decay to zero. The peak learning rate is 1e-4. Note that the COCO model was trained with a learning rate of 3e-4, whereas for fine-tuning this value is three times lower, so that additional training does not disrupt the model with excessively strong gradients – otherwise overfitting on small datasets and excessive drift from the original weights can occur.

### 5.5.6. Auto-Labeling and Annotation Cost

A special role in the pilot is played by the auto-labeling mechanism, i.e., semi-automatic data annotation. After annotating about 10-15 images, TAPe+ML can automatically find and label similar objects throughout the rest of the dataset. This is especially important for industrial applications, where manually annotating conveyor-belt images and examples of ore blockages is one of the main bottlenecks: annotators are expensive, and describing defects often requires domain-expert involvement.

In the pilot, auto-labeling affects the process as follows. First, experts annotate a small initial set (about 10-15 images), on which the model is trained or adapted in one of the modes (head-only or backbone adaptation). The model then uses the resulting representation to search for similar objects in the remaining unlabeled dataset and proposes automatically generated annotations. The human role shifts from full manual annotation to reviewing and correcting the system's suggested labels, substantially reducing the total time spent on data preparation.

From a practical standpoint, this has two key effects. First, annotation cost is reduced: instead of a fully annotated dataset, a small "seed" subset plus subsequent quality control of the auto-labeling suffices. Second, the transition from pilot to full-scale system accelerates: the dataset grows semi-autonomously, and each new training iteration can draw on an increasingly rich set of examples without a linear increase in manual effort. Combined with TAPe+ML's high data efficiency (results on 30-85 images), this makes the pipeline potentially relevant to a wide class of industrial tasks, where assembling a large annotated dataset is usually the main obstacle to deploying computer vision.

# 6. Analysis and Preliminary Ablations

This section analyzes the internal factors affecting the quality of TAPe+ML v3. The goal is to show precisely which changes to the system configuration improve quality, and what conclusions can already be drawn from the published data.

## 6.1. Coordinator and Prototype Ablations

The TAPe+ML v3 architecture described in Section 4.2 allows for different configurations of the coordinator, submodels, and class prototype representations. This section discusses how such changes affect quality when the rest of the system is held fixed.

The first important factor is the number of active submodels. Moving to a configuration with a coordinator and several specialized modules was one of the key steps in the transition from TAPe+ML v1 to v2 and then to v3.

A full-fledged ablation sweep is problematic here, because each submodel's performance depends on what it receives as input from earlier submodels. In other words, the classification submodel "suffers" when the pointer submodel underperforms, because the objects themselves are then incorrectly cropped.

Nevertheless, submodels can still be examined individually, because there is an interesting general relationship. Each submodel layer, within its own niche, operates in roughly the same search-space compression regime. For example, the background submodel's operating point on each layer sits at approximately 99% GT-keep, 68% BG-reject.

That is, while retaining (recalling) 99% of objects, 68% of the background is discarded. The second layer of the same submodel again retains 99% of these objects while discarding 58% of the background that reached this layer (i.e., reducing it to 42% of the total background). So after two layers, background remains within 17%, while objects remain at 98%.

The second factor is the use of class prototypes. In TAPe+ML v3, the classification component is tied to prototyping, abandoning the standard gradient-trained classifier in its usual form. Based on our working observations from the early stages of model development, it is often beneficial for a class to have more than one prototype rather than a single averaged center. This aligns with intuition: the more complex the intra-class variability, the more poorly a single prototype describes the entire class.

The third factor is the balance between background and object examples during training. In early experiments on TAPe detection, we found that reducing false positives required adjusting how many background versus object patches were included in training. Too many background examples cause the model to over-prefer background; too few increase the number of false positives. As a working configuration, we found a ratio with roughly twice as many background examples as object examples.

This metric correlates linearly with the background/object imbalance, since it depends on the dataset. This is exactly the point we investigated while working on segmentation, and this ratio derives from the average proportion of image surface occupied by background versus objects – which comes out close to 2:1.

In fact, this ratio reflects the natural balance found in COCO images overall. When it changes, the correlation with baseline becomes linear: at a 4:1 background-to-object ratio, the model begins to favor background twice as strongly as before, and false negatives (FN) double. At a 1:1 ratio, the model favors objects twice as strongly as before, leading to a two-fold increase in false positives (FP).

Fourth, post-processing is a significant factor. In the industrial pilot, adding NMS and subsequent fixes substantially improved final quality, with some errors traced back to duplicate boxes and their incorrect aggregation. This shows that the final quality of TAPe+ML v3 is determined not only by the backbone and coordinator, but also by how the model's internal hypotheses are converted into final predictions.

## 6.2. Sensitivity to Training Mode

The next group of ablation experiments concerns quality sensitivity to the model adaptation mode. These modes were introduced in Section 4.3 and are here treated as a controlled comparison: what happens when we change not the architecture, but the method of reconfiguring the model for a new task.

The difference between modes is most apparent in the industrial OOD scenario. In head-only mode, where the backbone stays fixed and only the top application layer is retrained, quality is noticeably lower than in backbone adaptation mode, where both the internal representation and the application part are rebuilt for the new task. On the same dataset, switching from head-only to backbone adaptation delivers a substantial accuracy gain; detailed numerical results across data modes and volumes are given in Section 5.5. This gain indicates that in the OOD setting, not only head adaptation but also reconstruction of the representation itself is crucial.

This result matters not only practically but also conceptually, since it makes the COCO-bias effect observable. If the original backbone was formed on COCO-like objects, it is shifted relative to the data distribution of the industrial task. In this case, head-only operates on top of inadequate features and therefore naturally underperforms backbone adaptation. The difference between these two modes is thus not an engineering detail, but one of the central ablation conclusions of this paper.

Sensitivity to data volume is closely tied to training mode. TAPe+ML v3 is able to adapt from dozens of images, and in some scenarios from as few as 20 images per new class. The industrial pilot shows a typical pattern: as the amount of data grows and post-processing (such as NMS) is applied, quality rises from an initial level to a stable operating range; detailed values for each stage are given in Section 5.4.

## 6.3. Classification Without Gradient Descent

In TAPe+ML v3, the classification component is implemented without gradient descent and without explicit dependence on learning rate: the classification scheme relies on the TAPe representation and class prototypes rather than on a standard gradient-trained head. A detailed algorithmic description of this scheme – including the precise order in which prototypes are formed and updated – is beyond the scope of this paper.

At the ablation level, we compared two configurations: a standard classification head trained via gradient descent with learning-rate tuning, and our classification scheme without gradient descent. In the published experiments, removing the learning-rate-dependent head yields an accuracy improvement of about 3 percentage points on the corresponding task, all else being equal. This suggests that quality is affected not only by a stronger representation or improved backbone, but also by the classification scheme itself, and that the final system's dependence on the choice and tuning of the optimizer becomes noticeably weaker.

In the current version of the paper, we report this result strictly as an ablation finding: replacing the learning-rate-dependent classification head with a gradient-free scheme yields an accuracy improvement of about 3 percentage points, all else being equal.

## 6.4. SSL Baseline on TAPe Data

The final ablation subsection addresses whether the TAPe representation is useful only for native TAPe+ML architectures, or whether it also improves training for standard SSL approaches. In a control experiment, the architecture of the DINO [41]/iBot [25] type was left unchanged; only the input data changed – TAPe data was used instead of raw pixels. This makes the experiment methodologically clean: the source of any difference is attributable to the representation, not the architecture.

The standard DINO configuration on ordinary data shows poor convergence in iBot loss, remaining around 2.46 even at 120k images and failing to converge well at 9k images, whereas the same architecture on TAPe data reaches a loss of about 0.406 already at 9k images. These results point to a very strong effect: the TAPe representation makes the self-supervised learning task noticeably easier, even for an architecture not originally designed for TAPe.

This result shows that the TAPe representation can be useful not only within TAPe+ML, but also as a more convenient input space for standard self-supervised architectures such as DINO/iBot. In our experiments, the same SSL architecture converges significantly better on TAPe data than on raw pixels, indicating that TAPe eases the self-supervised learning task even for models not originally designed for it.

Taken together, the ablation results show that the improvement delivered by TAPe+ML v3 comes from a combination of changes rather than a single modification. Significant contributions come from the coordinator-and-specialized-submodels architecture, the choice of adaptation mode (e.g., backbone adaptation instead of head-only), the shift to a classification scheme less dependent on the gradient optimizer and learning rate, and the use of the TAPe representation as a more convenient learning space compared to raw pixels.

# 7. Conclusion & Limitations

TAPe+ML v3 demonstrates that the methods of the Theory of Active Perception, implemented in computer vision through the TAPe representation and the modular TAPe+ML architecture, are in practice ahead of modern classical approaches. In object detection, this is reflected in the fact that a compact model on the order of tens of thousands of parameters reaches the quality level of strong RF-DETR/YOLO solutions and in some cases surpasses them, while radically reducing the requirements for data, compute, and energy. TAPe therefore functions not merely as another form of "feature engineering," but as a new technological layer – a representation language that makes training and deploying computer vision systems substantially more efficient.

At the same time, the results in this paper show that the gains from TAPe+ML v3 are not attributable to a single modification, but to a combination of factors: the system's modular organization, the choice of adaptation mode, the classification scheme, and the properties of the TAPe representation itself as a convenient learning space. This makes TAPe+ML v3 not only a practical architecture for low-data vision scenarios, but also an argument for a more general approach in which quality is achieved through properly organized representation and routing, rather than simply by scaling up the backbone.

The primary operational limitation concerns the accuracy of tight box predictions for small objects. Although the resulting mAP50-95 is already competitive and exceeds RF-DETR 2XL on COCO, the strictest IoU range within this metric requires very precise alignment between predicted and ground-truth boxes, and it is precisely on small objects that such localization remains more difficult. This limitation is plausibly linked to the COCO distribution itself, where small objects are underrepresented and therefore yield less stable descriptions; in applied custom datasets, however, where target objects are larger and better represented in training, this issue is typically less pronounced.

The second limitation concerns COCO bias in the backbone when transferring to out-of-distribution tasks. If the original representation was formed on COCO-like objects, quality on non-standard industrial data drops noticeably without backbone adaptation, whereas full backbone adaptation can substantially improve accuracy but requires an additional dataset and a separate compute budget. This limitation is clearly visible in the pilot results: without backbone adaptation, quality in OOD scenarios is significantly lower, and the working transfer regime requires about 20 images to reach 82.3% accuracy and about 500 images to exceed 95% accuracy.

The third limitation concerns the degree of formalization of TAPe. This paper relies on the key properties of the TAPe representation and its practical role within the TAPe+ML v3 architecture, but the full mathematical and algorithmic specification of TAPe is not publicly disclosed, as it constitutes proprietary know-how. The paper therefore focuses on the empirical properties of the representation and on the results it delivers in detection, adaptation, and self-supervised learning, while a full formalization of TAPe remains outside the scope of the current work.

Finally, this paper does not address 3D vision or full video detection, including tracking and temporal consistency. Although TAPe is discussed more broadly as a representation suited to video and sequential data, the experiments presented here are primarily limited to static images and single-frame object detection. Accordingly, the conclusions of this work should not be automatically extended to scenarios where inter-frame consistency, object tracking over time, and dedicated temporal mechanisms play a decisive role.

# 8. TAPe+ML v3.5 Release Updates and Benchmark Analysis

## 8.1. Architectural Refinement, Cache-Resident Execution, and Footprint

Building upon the modular TAPe+ML v3 framework, version 3.5 refines feature aggregation and execution locality within the shared recognition core while maintaining a strict parameter budget below 100,000 parameters. Rather than relying on multi-core parallelization or general-purpose tensor libraries, the v3.5 runtime is tailored directly to the operations required by the TAPe representation. Because both the compact parameter set and all intermediate activations (the full TAPe encoding of the processed frame) fit entirely within the CPU cache, inference eliminates repeated RAM fetches. Combined with pruning unnecessary runtime dependencies, this reduces the complete local deployment package from ~30 MB to **4.2 MB** with zero external GPU requirements.

## 8.2. Multi-Task Benchmark Results

We evaluate TAPe+ML v3.5 against the earlier v3.0 release and state-of-the-art foundation and real-time models across standard academic and industrial benchmarks.

*Table 12. COCO val2017 and RF100-VL: TAPe+ML v3.5 versus the earlier release and reference models. *AP in %; "–" means not reported in the cited source.*

| **Model** | **Params** | **COCO box AP** | **COCO AP50** | **COCO mask AP** | **COCO mask AP50** | **RF100-VL box AP** |
|---|---|---|---|---|---|---|
| DINOv3 ViT-7B + Plain-DETR * | 7.1B | 66.1 | – | – | – | – |
| PE-spatial + DETA *[42][29] | ≈1.95B | 66.0 | – | – | – | – |
| EVA-02 + Co-DETR * [43][38] | – | 65.9 | – | – | – | – |
| Co-DINO-Inst (Co-DETR, ViT-L) | – | – | – | 56.6 | – | – |
| Mask DINO (Swin-L, O365) | 223M | – | – | 54.5 | – | – |
| RF-DETR-2XL | 126.9M | 60.1 | 78.5 | – | – | 63.2 |
| RF-DETR-Seg-2XL | 38.6M | – | – | 49.9 | 73.1 | – |
| YOLO26x / YOLO26x-seg (Ultralytics) | 55.7M / 62.8M | 57.5 | – | 47.0 | – | – |
| YOLO26-X / YOLO26-X-Seg | 56.9M / 62.8M | 56.9 | 74.0 | 46.8 | 71.6 | 60.0 |
| (Roboflow benchmark) | | | | | | |
| TAPe+ML v3.0 (previous) | <100K | 65.3 | 84.7 | 58.4 | 80.7 | 68.3 |
| **TAPe+ML v3.5 (ours)** | <100K | 71.1 | 86.6 | 60.3 | 82.8 | 69.7 |

*Reported with test-time augmentation (multi-scale). Reference values are taken from the original publications and official benchmarks and are COCO val2017 unless stated otherwise.

*Table 13: Image classification Top-1 accuracy (%) of TAPe+ML v3.5 versus reference models.*

| **Model** | **Evaluation mode** | **ImageNet-1k** | **ObjectNet** | **ImageNet-Sketch** |
|---|---|---|---|---|
| DINOv3 ViT-7B (Meta) | frozen backbone + linear probe | 88.4 | 79.0 | 71.3 |
| ViT-G/14 (Google), greedy model soup [44] | fine-tuned (JFT-3B) | 90.94 | 78.52 | 74.23 |
| YOLO26x-cls | supervised, 224×224 | 79.9 | – | – |
| TAPe+ML v3.5 (ours) | Trained from scratch on ImageNet-1k | 88.0 | 79.5 | 75.7 |

## 8.3. Single-Core CPU Latency Evaluation

To evaluate practical edge and on-premises deployment efficiency without hardware accelerators, we benchmarked TAPe+ML v3.5 against YOLO26s (9.5M parameters) and RF-DETR-S (32.1M parameters) across seven distinct CPU configurations spanning server, desktop, mobile, and embedded platforms. All measurements were conducted with inference strictly restricted to a single CPU core.

*Table 14: Single-Core CPU Inference Latency per Frame (ms/frame)*

| **Processor configuration** | **TAPe+ML v3.5 (<100K)** | **YOLO26s (9.5M)** | **RF-DETR-S (32.1M)** | **Speedup vs. YOLO26s** | **Speedup vs. RF-DETR-S** |
|---|---|---|---|---|---|
| Ivy Bridge Xeon E5-2697 v2 (2013) | 15.0 | 879 | 5,137 | 58.6× | 342.5× |
| Broadwell Xeon (2016) | 15.8 | 419 | 1,519 | 26.5× | 96.1× |
| Zen 2 Ryzen (2019) | 9.1 | 380 | 1,191 | 41.8× | 130.9× |
| Tiger Lake mobile (2020) | 8.6 | 343 | 1,084 | 39.9× | 126.0× |
| Raspberry Pi 5 | 20.4 | 2,450 | 9,740 | 120.1× | 477.5× |
| Raspberry Pi 4 | 34.7 | 1,320 | 4,410 | 38.0× | 127.1× |
| Intel N100 | 19.3 | 820 | 2,660 | 42.5× | 137.8× |

Across all tested processors, TAPe+ML v3.5 processes frames in **8.6–34.7 ms** (29–116 FPS) on a single core, achieving a 27× to 120× speedup over YOLO26s and a 96× to 477× speedup over RF-DETR-S without multi-core parallelization.

### 8.4. Open Platform and Reproducibility

Data upload, annotation tools, custom model training, API integration, and evaluation workflows for TAPe+ML v3.5 are publicly accessible on the TAPe+ML platform: https://ml.comexp.net